\pdfoutput=1
\documentclass[11pt]{article}

\usepackage{ACL2023}

\usepackage{times}
\usepackage{latexsym}

\usepackage[T1]{fontenc}
\usepackage[utf8]{inputenc}

\usepackage{microtype}

\usepackage{inconsolata}

\usepackage{multirow}
\usepackage{multicol}

\usepackage{amsmath} %for math equation
\usepackage{amssymb} %for checkmark
\usepackage{algpseudocode} %for algorithms
\usepackage{algorithm}
\usepackage{float}
\usepackage{graphicx}
\usepackage{booktabs}

\usepackage{ulem}

\title{\textsc{FlashBack} : \\ Efficient Retrieval-Augmented Language Modeling for Fast Inference}

\author{Runheng Liu$^{\dagger}$,~~Xingchen Xiao$^{\dagger}$,~~Heyan Huang\thanks{\ \ Corresponding author.},~~Zewen Chi,~~Zhijing Wu\\
School of Computer Science and Technology, Beijing Institute of Technology\\
\texttt{\{rhliu,xcxiao,hhy63,czw,zhijingwu\}@bit.edu.cn}\\}

\begin{document}
\maketitle
\def\thefootnote{${\dagger}$}\footnotetext{These authors contributed equally to this work.}
\def\thefootnote{\arabic{footnote}}

\begin{abstract}
  Retrieval-Augmented Language Modeling (RALM) by integrating large language models (LLM) with relevant documents from an external corpus is a proven methodology for enabling the LLM to generate information beyond the scope of its pre-training corpus. Previous work by retrieving a set of tokens iteratively with retrieved content prepending to the input poses a high run-time issue, which degrades the inference efficiency of the LLMs because they fail to use the Key-Value (KV) cache efficiently. We propose \textsc{FlashBack}, a modular RALM designed to improve the inference efficiency of RALM with the appending context pattern while maintaining decent performance after fine-tuning by Low-Rank Adaption. \textsc{FlashBack} appends retrieved documents at the end of the context to efficiently utilize the KV cache. We also introduce the Marking Token as two special prompt tokens for marking the appending context during fine-tuning. Our experiments show that \textsc{FlashBack} can improve language modeling performance in the perplexity metric. We proved that the Marking Token is a usable add-on when fine-tuning models on specific context patterns. By bypassing unnecessary recomputation, \textsc{FlashBack} achieves fast inference speed with long context input. The inference speed is up to $4\times$ faster than the prepending counterpart on a 7B LLM (Llama 2) in the runtime test.

\end{abstract}
\footnote{Our code on GitHub: \newline \url{https://github.com/BIT-NLP-GROUP/FlashBack}}
\section{Introduction}
\textit{}\begin{figure}[t]
\centering
\hspace*{-11pt}
\includegraphics[width=1.06\columnwidth]{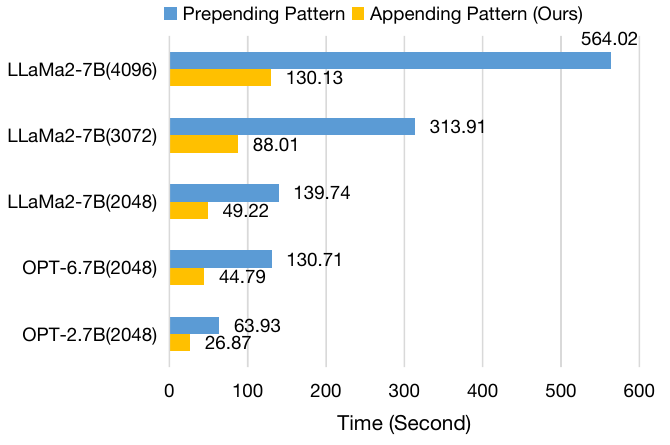}
\vspace{-15pt}
\caption{
A series of inference tests with simulated input and retrieved content on OPT and Llama 2 for comparing Appending Pattern with Prepending Pattern. (Maximum sequence length is denoted under the model name)
}
\label{fig:show}
\end{figure}
\begin{figure*}[t!]
\centering
\hspace*{-11pt}
\includegraphics[width=\textwidth]{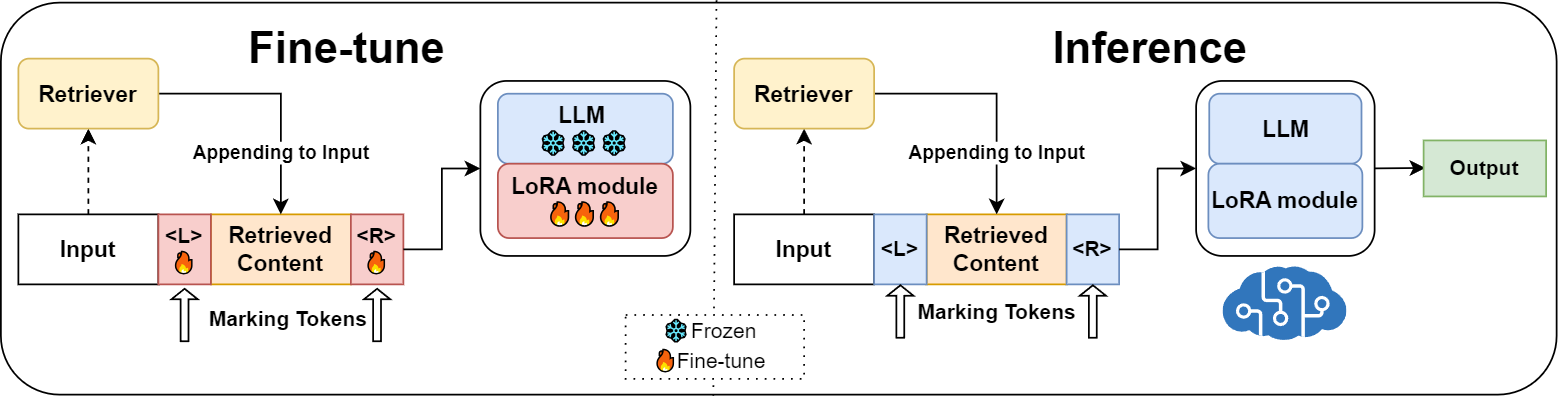}
\vspace{5pt}
\caption{
An illustration of the \textsc{FlashBack} pipelines. The left side of the diagram is the fine-tuning phase of our model with Marking Token and LoRA module. The inference procedure is on the right. 
}
\label{fig:pipeline}
\end{figure*}
Large language models (LLMs) based on the Transformer architecture \cite{vaswani2023attention} such as GPT, Llama, and OPT, etc. \cite{brown2020language,touvron2023llama,zhang2022opt} require enormous computational resources to keep their knowledge up-to-date \cite{meng2023locating}. Thus, Retrieval-Augmented Language Modeling (RALM) has emerged as a popular approach, enabling content generation that leverages external corpora to extend beyond the knowledge inherent in the model's parameters, thereby reducing the computational cost of capturing up-to-date knowledge. The effectiveness of using RALM to improve the performance of LLM has been demonstrated by previous studies \cite{guu2020realm,lewis2021retrievalaugmented, izacard2022atlas,wang2023shall}.

Recent works utilizing retrieved content simply by prepending retrieved contents to the input without updating language models exhibit the impressive potential of adapting off-the-shelf LLMs to the retrieved content \cite{shi2023replug,ram2023incontext}. Other studies build RALM through the pre-training of LLM \cite{borgeaud2022improving}, or engage in the continual pre-training of existing RALMs \cite{izacard2022atlas, wang2023instructretro}.

However, these works are introduced with limitations. First, the off-the-shelf LLMs are not inherently trained to incorporate retrieved content, and extensive pre-training of LLMs for building RALM incurs high computational costs \cite{lin2023radit}. 
Second, although in-context methods have been effectively applied on off-the-shelf LLMs \cite{ram2023incontext}, our work verified that prepending long input with In-Context RALM poses a high computational cost, which increases the inference time. To be specific, retrieved content changes with every retrieval stride tokens, and the computed key-value (KV) cache of the entire context is discarded, requiring re-computation for each change. The text generation process with a long input context can substantially increase the inference runtime, and the attention KV cache grows even larger on LLM with hundreds of billions of parameters \cite{pope2022efficiently}. Under these circumstances, recomputation of the KV cache becomes a computationally intensive task, entailing significant costs for RALM with continuous retrieval \cite{ram2023incontext}.

\begin{figure*}[t!]
\centering
\hspace*{-11pt}
\includegraphics[width=\textwidth]{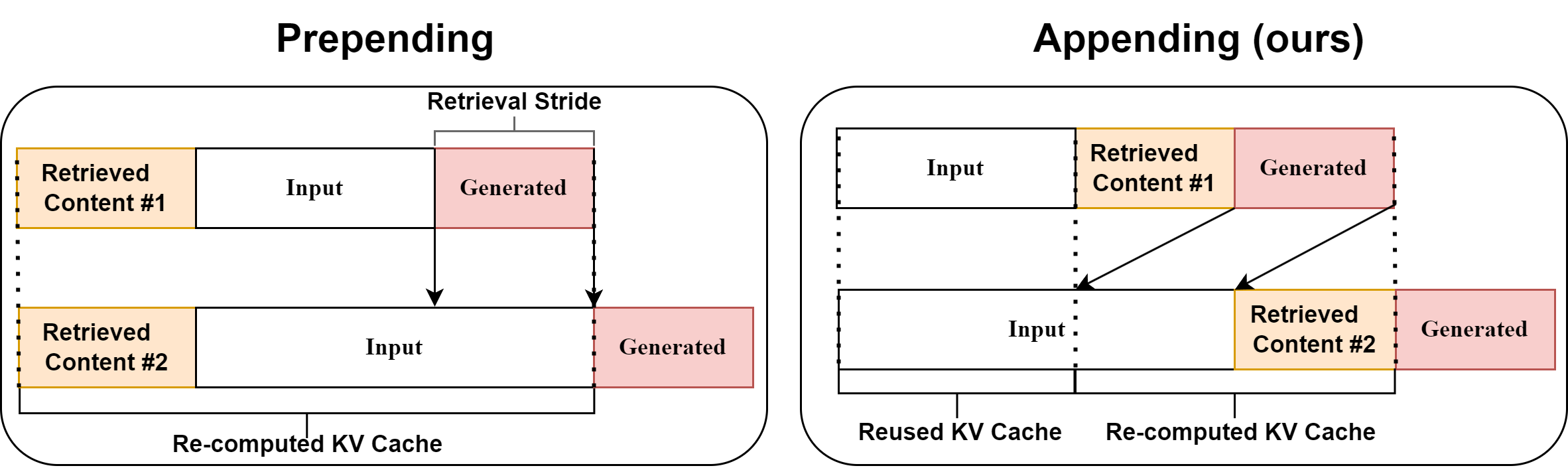}
\vspace{5pt}
\caption{
Comparison of prepending and appending context patterns in interval of change of retrieved content, arrow represents that generated tokens in past round will become a part of input in the next round.
In the prepending context pattern,the key-value of all tokens in past round should be re-computed. By contrast, in appending context pattern, the key-value of input in past round can be reused in the next round.
}
\label{fig:method}
\end{figure*}

In this paper, we propose \textsc{FlashBack} to overcome these issues we mentioned above, a method that utilizes off-the-shelf LLMs efficiently without further pretraining of LLMs as shown in Figure \ref{fig:pipeline}, to address the inefficiency problem by breaking the dependence of the input to retrieved content in the context. 
To be specific, instead of prepending the retrieved content directly on the input, we utilize the \textbf{Appending Context Pattern}, in which the retrieved content is appended at the end of the input (see Figure \ref{fig:method}). 
In our method, the KV cache of each input is static during inference, and the recomputation of the KV cache takes place only in the part of the retrieved content. 
Although there are advantages in applying this context pattern, it also presents an issue that degrades the performance of the RALM. 
Our appending pattern breaks the semantic coherence of the context, leading to a loss in perplexity when applying it to existing RALM (see Table \ref{tab:app_arxiv}). 
To solve this issue, inspired by \citet{ren2023context},  \textsc{FlashBack} introduces a tunable \textmd{Marking Token} to adapt to our appending context pattern for restoring comparable performance.

In particular, we use LoRA \cite{hu2021lora} to fine-tune the Marking Token while keeping both the retriever and the LLM frozen. Therefore, \textsc{FlashBack} is an adaptable approach that can cooperate with other RALM that perform retrieval for every $n$ token, thereby aiming for efficient fine-tuning, inference and context-model alignment. 

\mbox{}
\\
\\
Our main contributions can be listed as follows:

\begin{itemize} 
  \item Our work analyzed the prepending context in the FLOPs analysis and experiments. We find it is inefficient in RALM with long input due to redundant re-computation of the KV cache (see Figure \ref{fig:method} and section \ref{sec:3.3}). 

  \item We find that the \textbf{Appending Context Pattern} can improve the inference efficiency of RALM. (Up to $4\times$ faster in 7B Llama 2 as shown in Figure \ref{fig:show})

  \item We exploit \textbf{Marking Token} that can be used for adapting the LLMs to new context patterns by efficient fine-tuning with LoRA instead of pre-training.
\end{itemize}

\section{Background and Related Work}

\paragraph{Retrieve-Read RALM} 
Previous works \citep{borgeaud2022improving,lin2023radit,ram2023incontext,shi2023replug} have created distinct modules for document selection and document reading, and this methodology can be defined as the Retrieve-Read RALM. In the recent RALM framework, particularly for those employing LLMs, the imperative is to align the retrieved documents with the specific requirements of the LLMs \citep{gao2024retrievalaugmented}. AAR, REPLUG, and UPRISE are fine-tuning retrievers with frozen LLMs to achieve alignment \citep{cheng2023uprise,shi2023replug,yu2023augmentationadapted}. RETRO uses frozen retrievers to build their own RALM \citep{borgeaud2022improving}. In-Context RALM uses a frozen retriever for document selection and a frozen LLM for document reading without undergoing additional training for the LLM or the retriever \citep{ram2023incontext}. RAVEN \citep{huang2023raven} continually pre-train ATLAS \citep{izacard2022atlas} for improving its in-context capabilities, which aligns the model to specific prompting strategy, and they also proposed Fusion-in-Context Learning for incorporating more in-context examples during its inference process.

    Hence, our modeling is close to the RALM genre that uses decoder-only LLMs \textbf{(Auto-regressive Models)} and \textbf{Retrieval-Read RALM} that uses distinct modules for document selection and document reading \citep{borgeaud2022improving,ram2023incontext,lin2023radit,shi2023replug}. Furthermore, our proposed method is classified as \textbf{Input augmentation} by recent study \citep{asai2024reliable}. After all, this type of RALM has recently been defined as Modular RAG (Retrieval-Augmented Generation) and is increasingly becoming the dominant norm in the RAG domain \citep{gao2024retrievalaugmented}. Furthermore, it is worth mentioning that FinanceBench \citep{islam2023financebenchnewbenchmarkfinancial} also investigates the effectiveness of the context pattern. However, their efforts in modifying context positions are solely focused on inference accuracy rather than efficiency. GRIT-LM \citep{muennighoff2024generativerepresentationalinstructiontuning} improves model inference efficiency by reusing embedding-based representations that only apply to the embedding model. Our work \textsc{FlashBack}, is a text-based method that is not limited to embedding models.
\section{Methodology}
\label{headings}

\subsection{RALM with In-Context-Learning}\label{sec:icralm}
In the In-Context RALM framework \citep{ram2023incontext}, an external corpus $\mathcal{C}$ is provided to the retriever. The retriever is responsible for determining the specific content within the corpus to be utilized as the retrieved content, and then the retrieved content can be concatenated with input to form a context that conditions the predictions of the LLM. Given a sequence of tokens $x_1,...,x_n$ with length $n$, The probability of token sequence is represented by
\begin{equation}
    p(x_1,...,x_n) = \prod_{i=1}^n p_{\theta}(x_i|[\mathcal{R}_\mathcal{C}(x_{<i});x_{<i}]),
\end{equation}
where $\theta$ is parameters of an auto-regressive model, $\mathcal{R}_\mathcal{C}(x_{<i})$ is retrieved content from corpus $\mathcal{C}$ using prefix $x_{<i}$ and $[\mathcal{R}_\mathcal{C}(x_{<i});x_{<i}]$ denotes the concatenation of strings $\mathcal{R}_\mathcal{C}(x_{<i})$ and $x_{<i}$.

For the trade-off between runtime cost and model performance, the retrieval stride $s$ and query length $\ell$ are proposed to perform the retriever call less frequently(at least, every s > 1 token).
Therefore, the retrieval would be performed for every $s$ token, using the last $\ell$ tokens as designated retrieved content for input. 

The probability of the token sequence is represented by
\begin{multline}
p(x_1,...,x_n) = \\
 \prod_{j=0}^{n_s-1}\prod_{i=1}^s p_{\theta}(x_{s\cdot j+i}|[\mathcal{R}_\mathcal{C}(q_j^{(s,\ell)});x_{<s\cdot j+i}]),
\label{eq:prepending}
\end{multline}
where $n_s=n/s$ is the number of retrieval strides, $q_j^{(s,\ell)}=x_{s\cdot j-\ell+1},...,x_{s\cdot j}$ is the retriever input.

Our work followed this retrieval mechanism in addition to using different context patterns because this mechanism uses distinct retrieval modules and readers (LLMs) so that they can be optimized independently.

\subsection{Context Pattern}

In decoder-only transformer-based models, the computation of attention modules is related to the query of the current token and the key-value representations of preceding tokens \citep{vaswani2023attention}. 
During inference, the key-value of the current token would be saved in the KV cache, eliminating the requirement in the subsequent step.
We find that prepending retrieved content to the input has been prevalent in previous methods \citep{ram2023incontext,shi2023replug}. 
However, the key-value computed for previous tokens becomes obsolete, and a re-computation of the new key-value is required for each retrieval step since the retrieved content varies with each retrieval.

To avoid the extensive cost of re-computation, we remove the autoregressive dependency between retrieved content and input in context by using an appending context pattern. Specifically, we append retrieved content to the input (at the end of the input) as shown in Figure \ref{fig:method}.  

Then, the probability of the token sequence is represented by
\begin{multline}
p(x_1,...,x_n) = \\ \prod_{j=0}^{n_s-1}\prod_{i=1}^s p_{\theta}(x_{s\cdot j+i}|[x_{<s\cdot j+i};\mathcal{R}_\mathcal{C}(q_j^{(s,\ell)})]).
\label{eq:appending}
\end{multline}

In this case, for each input, the change of retrieved content $\mathcal{R}_\mathcal{C}(q_j^{(s,\ell)})$ would not require a re-computation of new value, therefore preserving the computed value of previous tokens in KV cache $x_{<s\cdot j+i}$.
(Algorithm pseudo code in Appendix \hyperref[algo:ficralm]{A}.)

\subsection{FLOPs analysis of context pattern}
\label{sec:3.3}
We analyzed the Floating Point Operations (FLOPs) of recomputation.
In the following analysis, the input batch size is denoted as $b$, the hidden size of the model as $h$, and the dimensions of the Key and Value vectors are also $h$, with a maximum sequence length of $T$, the retrieval stride of $s$, and a model composed of $l$ layers, then the size of the input vector is given by $[b, i \cdot s, h]$. The dimensions of the key and value projection matrix are $[h, h]$, and the number of retrievals is denoted as $i$. Overall, the FLOPs of re-computation are formulated as:
\begin{equation}
    C_0=2l\sum_{i=1}^{T/s}2bish^2=\frac{2T(T+s)bh^2l}{s}.
\label{eq:quadratic}
\end{equation}
It increases quadratically as a function of the maximum sequence length $T$, and thus, it will increase dramatically when the maximum sequence length $T$ is at a large value.

\subsection{Marking Token and Fine-tuning Choice}
Since LLMs are not aligned explicitly with our appending pattern, we use the Marking Token and Low-Rank Adaption (LoRA) \citep{hu2021lora} to adapt them to the appending pattern while keeping the origin model weights frozen so that alignment is achieved without modifying the inherent ability of the LLMs. \textmd{Marking Tokens} are represented as two special prompt tokens denoting \textmd{<MARK\_L>,<MARK\_R>} which are added into the vocabulary of the model, and they are used for marking the boundary of the retrieved content in context (see Figure \ref{fig:pipeline}). Then, we fine-tune the model using LoRA with Marking Tokens applied for adapting the appending pattern. LoRA is a parameter-efficient fine-tuning (PEFT) for fine-tuning a subset of parameters in the pre-trained model while targeting for obtaining performance comparable to full model fine-tuning. Recent studies suggest that forgetting is inevitable in fine-tuning, but the PEFT method can facilitate less forgetting during the fine-tuning process, which does not greatly damage the performance of the pre-trained LLM \citep{kalajdzievski2024scaling}. In \textsc{FlashBack}, we opt for LoRA to demonstrate how fine-tuning a relatively minimal set of model weights enables our model to adapt to the appending context pattern while keeping pre-training model weights frozen.
Our experiments show that \textsc{FlashBack} has significantly increased inference speed and can still maintain comparable perplexity (PPL) after fine-tuning. During fine-tuning, all parameters are frozen except for the embedding of <MARK\_L>, <MARL\_R>, and LoRA modules applied to attention layers.
\subsection{FLOPs analysis of Marking Token and LoRA}
If the rank of LoRA weight matrices is $r$, then the dimensions of the projection matrix are $[h,r]$. The average length of the retrieved content is $d$. We only equip LoRA modules, two projection matrices, for Key and Value matrices. Overall, the FLOPs of LoRA modules in appending context pattern can be formulated as:\footnote{We only consider multiplication operations and omit some addition operations.}
\begin{multline}
    C_1  = 2l(4bThr + bTh) \\ +2l\sum_{i=1}^{T/s}\left[4b(d+s)hr+b(d+s)h\right] 
    \\  = \frac{2l(4r+1)bhT(d+2s)}{s}   
\label{eq:LoRA}
\end{multline}
Combining equation \ref{eq:quadratic} and \ref{eq:LoRA}, the decreament of FLOPs when using appending context pattern with LoRA is:
\begin{multline}
    C_{\text{decreament}} = C_0-C_1 
    \\ = \frac{2lTbh\left[(T+s)h-(4r+1)(d+2s)\right]}{s}              
\label{eq:appending+LoRA}
\end{multline}

\subsection{Retriever}
Our runtime test used a sparse model, BM25 \citep{bm25}, and the QA task is tested with the DPR retriever \citep{karpukhin2020densepassageretrievalopendomain} to demonstrate our idea. In practice, \textsc{FlashBack} intrinsically supports switching to other retrievers in a plug-and-play manner.

\section{Experiments}

\subsection{Run-time Improvement}

\paragraph{Setup}
We test OPT, GPT-2, and Llama2 in different model sizes and employ simulated retrieved content and input in inference speed tests. Our experiment included tests with varying input lengths. In addition, it involved examining the inference time under two distinct context patterns.

\begin{figure*}[t!]
\centering
\hspace*{-11pt}

\includegraphics[width=\textwidth]{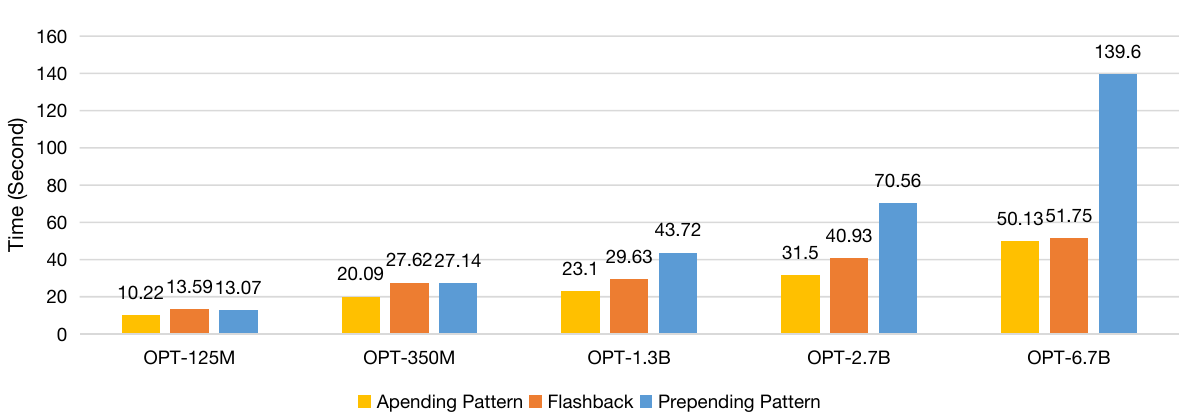}
\vspace{-15pt}
\caption{
Time for OPT models generating a sequence of length 2048 tokens (including 128 tokens for retrieved content), given a input of 512 tokens, with retrieval stride of 16.
}
\label{fig:opt}
\end{figure*}

\begin{figure}[h!]
\centering
\hspace*{-5pt}
\includegraphics[width=1.06\columnwidth]{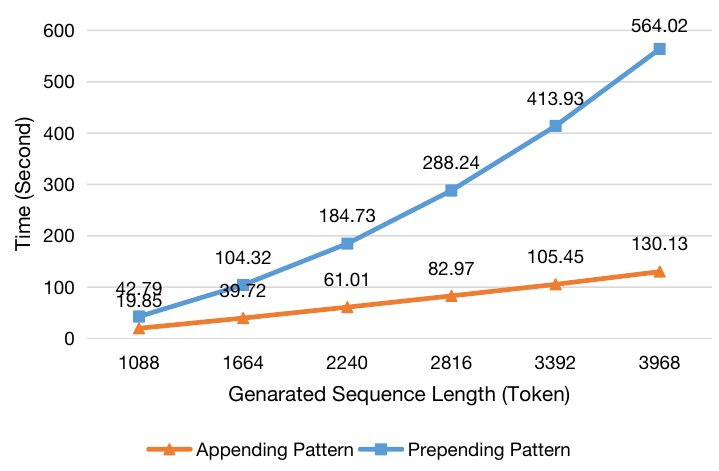}
\vspace{-15pt}
\caption{
Time for Llama2-7B generating sequences with different length, given a input of 512 tokens, with retrieval stride of 16.
}
\label{fig:llama2}
\end{figure}

\paragraph{Results}
In Figure \ref{fig:opt}, we compare the appending pattern (in-context), prepending pattern (in-context), and \textbf{FlashBack} in OPT models. 
We scale the model size from 125M to 6.7B.
Acceleration of inference time is more effective in large models that have more layers and larger hidden sizes.
In Figure \ref{fig:llama2}, we scale the maximum sequence length from 1088 to 3968 and observe a significant improvement, since Llama 2 has a larger maximum sequence length, which is 4096 tokens. (Note: The test results for inference speed may vary depending on the hardware used and setting; our runtime tests are conducted on a single A100-80GB GPU setting to P0 state)

\subsection{Language Modeling}\label{sec:4.2lm}

\begin{table*}[h!]
\centering
\begin{tabular}{cccccccc}
\toprule
Model &A./P. & LoRA & M.T. & Wikitext-2 & Arxiv & Freelaw & Stackexchange\\

\midrule

\multirow{7}{*}{OPT-1.3B}
&N.  &          &           &16.76&9.64&8.58&7.78\\
&P.  &          &           &15.02&9.59&8.35&7.63\\
&P.  &\checkmark &           &10.23&8.33&7.47&6.87\\
&P.  &\checkmark &\checkmark &\textbf{10.22}&\textbf{8.31}&\textbf{7.45}&\textbf{6.87}\\
%\cline{2-8}
&A.  &          &           &81.57&51.94&53.73&42.99\\
&A.  &\checkmark&           &12.65&11.38&10.11&9.13\\
&A.  &\checkmark&\checkmark &\textbf{10.49}&\textbf{8.72}&\textbf{7.85}&\textbf{7.17}\\
\midrule
\multirow{7}{*}{OPT-2.7B} 
&N.  &          &           &14.50&8.72&7.74&6.99\\
&P.  &          &           &13.14&8.68&7.56&6.88\\
&P.  &\checkmark&           &9.23&7.69&6.80&6.23\\
&P.  &\checkmark&\checkmark &\textbf{9.23}&\textbf{7.68}&\textbf{6.78}&\textbf{6.22}\\
%\cline{2-8}
&A.  &          &           &76.41&50.68&51.78&42.48\\
&A.  &\checkmark&           &11.58&10.88&9.31&8.51\\
&A.  &\checkmark&\checkmark &\textbf{9.52}&\textbf{8.08}&\textbf{7.19}&\textbf{6.53}\\
\midrule
\multirow{7}{*}{OPT-6.7B} 
&N.  &          &           &12.30&7.74&6.94&6.22\\
&P.  &          &           &11.20&7.73&6.83&6.15\\
&P.  &\checkmark&           &\textbf{8.23}&\textbf{6.98}&6.19&5.58\\
&P.  &\checkmark&\checkmark &8.24&6.99&\textbf{6.18}&\textbf{5.58}\\
%\cline{2-8}
&A.  &          &           &68.31&46.53&48.33&40.25\\
&A.  &\checkmark&           &10.54&10.92&9.27&8.04\\
&A.  &\checkmark&\checkmark &\textbf{8.59}&\textbf{7.43}&\textbf{6.64}&\textbf{5.94}\\
\bottomrule
\end{tabular}
\caption{
Perplexity (the lower the better) of OPT models on the entire validation set of WikiText-2 dataset and partial test set of Arxiv, Freelaw and Stackexchange datasets. The retrieval stride is set to 16. \textbf{A.} and \textbf{P.} refers to the appending and prepending context pattern respectively. \textbf{M.T.} refer to the Marking Tokens. And \textbf{N.} refers to none retrieval.
}
\label{tab:app_arxiv}
\end{table*}

\paragraph{Setup}

We test \textsc{FlashBack} on Wikitext-2 \citep{merity2016pointer} dataset and three datasets: Arxiv, Freelaw, and Stackexchange from the Pile \citep{gao2020pile}. We report token-level average perplexity as the evaluation metric (excluding \textmd{<MARK\_L>,<MARK\_R>}). 

Our experiments use models of OPT (1.3B-6.7B) and GPT-2 (124M-1.5B).

We compare \textsc{FlashBack} with the appending context pattern without fine-tuning and also the prepending context pattern. (For smaller models that are below 774M parameters, the results are in Appendix C Table \ref{tab:app_arxiv_small})

\paragraph{Results}
Table \ref{tab:app_arxiv} shows the results of different configurations in the dev set. We can see that the appending context pattern applied directly on LLMs through the in-context method generates higher perplexity compared to the model using the prepending context pattern. Our result on fine-tuning with appending context patterns by using LoRA demonstrates that the model can adapt to such patterns after fine-tuning. Notably, with both the Marking Token and LoRA applied, the difference of perplexity results between two distinct context patterns becomes progressively alleviated as the size of the test model increases. We also test GPT-2 from 124M to 1.5B to validate this observation. (See Table \ref{tab:app_arxiv_small} in \ref{appendix:d}). Furthermore, we test Llama 3.2-3B to validate our language modeling in one of the newest open source models (See Table \ref{tab:llama3} in \ref{appendix:e}). 

\subsection{Language Modeling with Continuous Retrieval}

\paragraph{Setup} As the extra run-time cost of LoRA in comparison to the full-parameter is acceptable, we employ it in the following experiments. We continue to test \textsc{FlashBack} on the Wikitext-2 dataset \citep{merity2016pointer}  and design a multiple retrieval scenario. Specifically, given a token chunk $x_{1,...,n_c}$, we split it by retrieval stride to obtain prefixes: $x_{1,...,s},x_{1,...,2s},...,x_{1,...,n_c-s}$, where $n_c$ is the length of the chunk, and $s$ is the retrieval stride. Then we use each prefix with the corresponding retrieved passages to predict the next $s$ tokens. We report token-level average perplexity as the evaluation metric (excluding \textmd{<MARK\_L>,<MARK\_R>}).

\paragraph{Results} Like the results on the language modeling task, a similar trend is observed in Table \ref{tab:lm_new}. Both Marking Token and LoRA are necessary to improve model performance. The performance gap between context patterns decreases as the model size increases.

\begin{table}[h!]
\centering
\begin{tabular}{ccccc}
\toprule
Model & A./P. & LoRA & M.T. & Perplexity$\downarrow$ \\
\midrule
\multirow{5}{*}{OPT-1.3B}   &P. &  &  &26.30\\
   &P. &\checkmark &&14.82 \\
   &P. &\checkmark  &\checkmark  &14.79\\
   &A. &\checkmark &&15.41 \\
   &A. &\checkmark  &\checkmark  &15.20\\
%16.42(1k) -> 15.21(5k)
\midrule
\multirow{5}{*}{OPT-2.7B}   &P. &  &  &22.73\\
   &P. &\checkmark  &  &13.14\\
   &P. &\checkmark  &\checkmark  &13.14\\
   &A. &\checkmark  &  &13.72\\
   &A. &\checkmark  &\checkmark  &13.49\\
\midrule
\multirow{5}{*}{OPT-6.7B}   &P. &  &  &19.51\\
   &P. &\checkmark &&11.65 \\
   &P. &\checkmark  &\checkmark  &11.68\\
   &A. &\checkmark &&12.18 \\
   &A. &\checkmark  &\checkmark  &11.98\\

\bottomrule
\end{tabular}
\caption{
Perplexity of OPT models on the language modeling with continuous retrieval task, evaluated on the validation set of WikiText-2. The retrieval stride is set to 16.
}
\label{tab:lm_new}
\end{table}

\subsection{Question Answering}
As improvement to the perplexity of the model results in performance improvement in downstream tasks, perplexity is a proxy metric for improved general-purpose abilities. In the In-Context RALM, retrieved passages boost model performance both on upstream and downstream tasks--language modeling and question answering, with perplexity and EM score as metrics, respectively. Although FlashBack can improve models on language modeling tasks, it is unclear whether it affects the downstream ability of models.

\paragraph{Setup} We test FlashBack in different settings (see Table \ref{tab:qa}) in Natural Questions (NQ) \citep{10.1162/tacl_a_00276} and TriviaQA\citep{joshi-etal-2017-triviaqa} open-domain question answering datasets to observe the effect of context pattern and the marking token on the QA task. 

\paragraph{Results} Our results show that the context pattern does not evidently affect the model performance in the QA tasks after fine-tuning the model. However, the Marking Token can improve the model's performance in QA tasks, especially for the test with the appending context pattern.

\begin{table}[h!]
\centering
\begin{tabular}{ccccc}
\toprule
\multirow{2}{*}{A./P.}&\multirow{2}{*}{LoRA}&\multirow{2}{*}{M.T.}& \multicolumn{2}{c}{EM$\uparrow$} \\
&&&NQ & TriviaQA  \\
\midrule
P. & \checkmark & &17.0&43.3 \\
P. & \checkmark & \checkmark & 17.5 & \textbf{44.7} \\
A. & \checkmark & &18.7&42.5 \\
A. & \checkmark & \checkmark & \textbf{20.3} & 43.0 \\

\bottomrule
\end{tabular}
\caption{ QA Results of NQ, TriviaQA test set for OPT-6.7B. The model is fine-tuned on the language modeling with continuous retrieval task. We test with DPR retriever and keep top-2 retrieved documents.}
\label{tab:qa}
\end{table}

\subsection{Implementation Details}
\label{sec:implement}
\paragraph{Run-time Improvement}
We randomly sample simulated token sequences from a discrete uniform distribution $U\{500, 1000\}$ to become pseudo-input and pseudo-retrieved content. 
We maintain the length of simulated retrieved content to a fixed value of 128, then test models with different maximum sequence lengths. (Setup details are shown in the Appendix Table \ref{tab:hp_runtime}). 
For the retrieval stride, all models are set to 16.
All run-time experiments are performed using a single NVIDIA A100 GPU.

\paragraph{Language Modeling}

For every $s$ token in the train set (refer to target), we sample from a discrete uniform distribution $U\{T/2 ,T-s\}$ to obtain the start position of the target, where $s$ is the retrieval stride and $T$ is the maximum sequence length. For Wikitext-2, Wikitext-103, a superset of Wikitext-2, is used as a retrieval corpus. To avoid data leakage, target outputs are included in retrieved content, and we filter retrieved content with titles that match Wikitext-2 during data construction. 
Due to the much larger size of the Pile dataset, for Arxiv, Freelaw, and Stackexchange, the validation split and 80\% of the test split are a mixture as a retrieval corpus. The rest of 20\% of the test split is divided into two parts equally as the training and the test dataset, respectively.
We employ BM25 retrieval and configure the query length to 16, indicating the use of the preceding 32 tokens of the target as input for the retriever. We use the next token prediction objective for language modeling experiments. The loss is computed on target tokens in context.

\paragraph{Training Details}
The trainable parameters for all models include two token embeddings of \textmd{<MARK\_L>} and \textmd{<MARK\_R>}, and LoRA modules applied to all attention layers. The rank of LoRA weight matrices is set to 16. The batch size is set to 16. 
The learning rates for the GPT-2 models (124M-1.5B) are specified as [4e-4, 2e-4, 1e-4, 5e-5]. Correspondingly, the learning rates for the OPT models (1.3B-6.7B) are specified as [1e-4, 5e-5, 5e-5]. We train for 5K additional steps, with 10\% warm-up steps and a cosine learning rate schedule. For running on a single 24GB RAM GPU, it is feasible to use the bf16 format on the OPT-6.7B model and set the maximum sequence length to 512.

In our implementation, we use the Transformers library \citep{wolf-etal-2020-transformers} and BM25 retrieval from the Pyserini library \citep{10.1145/3404835.3463238}. All fine-tuning experiments are conducted by using 4$\times$24GB RAM GPUs.

\section{Ablation Study}
\paragraph{Marking Token}We have already tested the model using an appending context pattern that is solely fine-tuned with LoRA without Marking Token to compare with \textsc{FlashBack}. In Table \ref{tab:app_arxiv}, the test results of models with Marking Tokens and LoRA fine-tuning have lower PPL, which means that fine-tuning the Marking Token could further boost the performance of the RALM, and therefore meet our expectations. Furthermore, we also tested whether the Marking Token and LoRA modules add additional inference costs to our modeling. Our results in Figure \ref{fig:opt} show that \textsc{FlashBack} is slightly slower than the model without these modules, but the difference is small enough to be ignored compared to the overall increase in the inference speed-up. In the QA test (Table \ref{tab:qa}), the results show that the marking token can be effective in predicting the model performance in both context patterns.
\begin{table}[h!]
\centering
\begin{tabular}{cccc}
\toprule
\multicolumn{4}{c}{Perplexity} \\
\multirow{2}{*}{Model} & \multicolumn{3}{c}{Retrieval Stride} \\ \cmidrule(lr){2-4}
&16&32&64 \\
\midrule
GPT2-124M&28.11&23.64&21.80 \\
GPT2-355M&16.46&15.66&15.80 \\
GPT2-774M&13.46&13.12&13.24 \\
GPT2-1.5B&12.59&12.28&12.24 \\
\midrule
OPT-125M&24.73&22.42&21.26 \\
OPT-350M&17.29&16.71&16.71 \\
OPT-1.3B&11.51&11.29&11.31\\
OPT-2.7B&10.70&10.38&10.23\\
OPT-6.7B&9.40&9.10&9.09\\

\bottomrule
\end{tabular}
\caption{Perplexity (the lower the better) of GPT models and OPT models on the validation set of WikiText-2 dataset, varying retrieval stride, by using \textsc{FlashBack}.}
\label{tab:stride_ppl}
\end{table}
\paragraph{Retrieval Stride} Allocating the number of retrieval strides for each iteration is a design choice that is related to the performance of the model in both perplexity and inference time. As we mentioned above in section \ref{sec:icralm},  the In-Context RALM \citep{ram2023incontext} chooses to use a cherry-picked retrieval stride to balance the performance trade-off. They choose to use a relatively small stride value to have better generation (lower perplexity) without adding extensive inference cost. However, our experiments (see Table \ref{tab:stride_ppl} and \ref{tab:stride_time} in the appendix) on varying retrieval strides on \textsc{FlashBack} implied a different opinion. We find that using 32 or 64 as retrieval strides does not degrade the model's performance. When testing with a larger model such as OPT-6.7B, the test with a higher retrieval stride has the lowest perplexity compared to the test using the same model with a lower retrieval stride. Therefore, the trade-off between perplexity and inference time may not be a particularly evident phenomenon for some cases. As a consequence, we made bold speculation that it might be an advantage for \textsc{FlashBack} to increase the retrieval stride without intensively degrading the model's perplexity and still maintaining high-speed inference in long context input scenarios. This phenomenon may require further demonstration in the future.
\section{Conclusion}
Prepending retrieved contents to the input without modification of model architecture has been a practicable method for LM to use external knowledge. However, the prepending context pattern hinders RALM from efficiently generalizing at a considerable context length with a long input.

This paper presented the \textsc{FlashBack} method, which enables RALM with faster inference speed while maintaining competitive model performance.
We proved that the appending context pattern is more efficient than the prepending counterpart. In addition, our experiments demonstrated that LLMs can adapt to new context patterns using Marking Tokens with adaption fine-tuning (LoRA). Its performance in generation quality is closely equivalent to the LLM using a prepending counterpart as the size of the model increases.

\section*{Limitations}

FlashBack is specifically designed for RALM methods that use multiple retrievals by every $n$ token. Other works using retrieval strategies like retrieval once or retrieval by every $n$ text chunk are not covered in our paper and require further study. In our experiments, the retrieval stride is a fixed value in each test, but it can be a dynamic variable, so there might be additional improvements. In the run-time comparison experiment, due to constrained computational resources, we used simulated (pseudo) data for testing. This is because we did not find an appropriate public downstream task that fulfills the following requirements: 1) long context input, 2) multiple retrieval, and 3) generated text becomes a new part of the context. QA task generally comes with a short question, which is a short input, so the burden of RALM mainly lies on the retrieved documents, rather than the input. However, we believe that this type of scenario exists in some applications where long input is required with multiple retrievals. Additionally, we did not test our modeling on LLMs over the size of 7B parameters since they generally require extensive computational resources.
 
Furthermore, in \textsc{FlashBack}, we did not scale the number of retrieved contents to a large value. It is practicable to generalize appending context patterns to more retrieved contents (for example, using the REPLUG framework\citep{shi2023replug}).

\section*{Ethics Statement}
This research explores the Retrieval-Augmented Generation (RAG) application leveraging Large Language Models (LLM). We call this method as Retrieval-Augmented Language Modeling (RALM).

Data in this research were obtained from publicly available datasets and sources that adhere to ethical guidelines, ensuring respect for user privacy and consent. Potential biases may exist inherently in both the external corpus and the LLMs. Using \textsc{FlashBack} for real applications requires additional precautions to avoid reinforcing stereotypes or propagating misinformation through the generation process.

\bibliography{my}
\bibliographystyle{acl_natbib}

\clearpage
\newpage
% \clearpage
% \hbox{}
% \newpage
\appendix
\section*{Appendix}

\section{FlashBack Algorithm}

\label{sec:appendix}

% This is a section in the appendix.
\begin{algorithm}[h!]

\caption{In-Context RALM with \textsc{FlashBack}}
\label{algo:ficralm}
Given retrieve stride $s$, retrieved query length $\ell$, and maximum sequence length $T$.

Given auto-regressive model $p_\theta$, retriever $q$, and initial prompt sequence $x_0,...,x_t$.

Initialise: $n \leftarrow t$, retrieved evidence $e \gets \text{None}$, Key-Value Cache $K\gets \text{None}$.

\begin{algorithmic}
\While {$n \leq T$}
    \State The retriever is called for every stride $s$ tokens.
    \If{$(n - t \mod s) = 0$}
        \State $e \gets q(x_{n-\ell,..,n})$ \Comment{Call retriever}
        \State $l_e \gets \text{length of }e$
        \State Merge past generated tokens to the context, using truncated KV cache:
        \State $p_n, K_{1,...,l_e+n-1} \gets p_{\theta}(x|[x_{1,...,\tilde{n}-1};x_{\tilde{n},...,n-1};e];K_{1,...,\tilde{n}-1})$
        \State $\tilde n \gets n$ \Comment{Save last retriever call position}
    \Else
        \State $p_n, K_{1,..,l_e+n-1} \gets p_{\theta}(x|[x_{1,..,\tilde{n}-1};e;x_{\tilde n,...,n-1}];K_{1,..,{l_e+n-2}})$
    \EndIf
    
    \State $x_n\sim p_n(x)$
    \State $n\gets n+1$
\EndWhile
\end{algorithmic}

\end{algorithm}

\section{Experiment Setup}
\begin{table}[h]
\centering
\begin{tabular}{ccc}
\toprule
Models      & Length of input   & Length limitation\\
\midrule
GPT2        & 256               &1024   \\
OPT         & 512               &2048   \\
Llama2      & 512               &4096   \\
\bottomrule
\end{tabular}
\caption{\label{tab:hp_runtime}
The length of input and the length limitation we used in run-time improvement experiments.
}
\end{table}

\begin{table}[h!]
\centering
\begin{tabular}{ccccc}
\toprule
\multicolumn{4}{c}{Time (Second)} \\
\multirow{2}{*}{Model} & \multicolumn{3}{c}{Retrieval Stride} \\ \cmidrule(lr){2-4}
&16&32&64 \\
\midrule
GPT2-124M&6.16&5.67&5.89 \\
GPT2-355M&10.68&10.37&10.35\\
GPT2-774M&16.08&15.97&15.77\\
GPT2-1.5B&21.71&21.23&20.26\\
\midrule
OPT-125M&10.72&10.09&10.10 \\
OPT-350M&20.01&19.9&19.69 \\
OPT-1.3B&22.13&20.72&19.79\\
OPT-2.7B&30.71&29.10&28.17\\
OPT-6.7B&50.25&43.79&41.06\\

\bottomrule
\end{tabular}
\caption{Time for GPT models and OPT models generating sequences with respective maximum context length, varying retrieval stride.}
\label{tab:stride_time}
\end{table}

\section{Language Modeling With More Retrieved Content}
Inspired by REPLUG \citep{shi2023replug}, we append each retrieved content separately to the input and ensemble respective output probabilities.
\paragraph{Setup}
We use the same setup as in Section \ref{sec:4.2lm}, except for varying the number of retrieved content.
\begin{table}[h!]
\centering
\begin{tabular}{ccccc}
\toprule
\multicolumn{4}{c}{Perplexity} \\
\multirow{2}{*}{Model} & \multicolumn{4}{c}{\# Of Retrieved Content} \\ \cmidrule(lr){2-5}
&1&2&4&8 \\
\midrule
GPT2-124M&25.00  &24.06   &24.65&25.72\\
GPT2-355M&16.05  &15.58   &15.91&16.50\\
GPT2-774M&13.28  &12.94   &13.21&13.67\\
GPT2-1.5B&12.49  &12.19   &12.43     &12.83\\
\midrule
OPT-125M&23.44  &22.53   &23.09&23.97\\
OPT-350M&16.89  &16.31   &16.67&17.29\\
OPT-1.3B&11.45  &11.10   &11.32&11.67\\
OPT-2.7B&10.48  &10.14   &10.32&10.63\\
OPT-6.7B&9.34  &9.05   &9.20&9.34\\

\bottomrule
\end{tabular}
\caption{Perplexity (the lower the better) of GPT models and OPT models on the validation set of WikiText-2 dataset, varying the number of retrieved content, by using \textsc{FlashBack}.}
\label{tab:multi_k}
\end{table}
\paragraph{Results}
From the test results of varying the number of retrieved contents, the perplexity of all models we tested slightly worsens as the number of retrieved contents increases. We find that the best perplexity comes with setting the number of retrieved contents to 2. We speculate that the reasoning behind this phenomenon is that the second retrieved document is likely to be highly relevant to the first retrieved document. However, this experiment is only for demonstrating that \textsc{FlashBack} can be used for multi-document retrieval. Further improvement on multi-document retrieval is not our focus in this paper, and it can be a direction in later work.

\section{Perplexity Results of GPT2 Model}
\label{appendix:d}
We tested the GPT2 models on various model sizes in perplexity metrics. The results are presented in Table \ref{tab:app_arxiv_small}.

\newpage
\begin{table*}[h!]
\centering
\begin{tabular}{cccccccc}
\toprule
Model &A./P. & LoRA & M.T. & Wikitext-2 & Arxiv & Freelaw & Stackexchange\\
\midrule
\multirow{7}{*}{GPT2-124M}
&N   &          &           &30.69&15.19&16.65&14.76\\
&P.  &          &           &26.00&14.98&16.00&14.43\\
&P.  &\checkmark&           &17.46&11.46&11.55&9.02\\
&P.  &\checkmark&\checkmark &17.41&11.44&11.53&9.00\\
&A.  &          &           &72.29&46.59&56.32&50.71\\
&A.  &\checkmark&           &25.56&22.35&24.31&17.41\\
&A.  &\checkmark&\checkmark &19.14&14.21&15.01&10.25\\
\midrule
\multirow{7}{*}{GPT2-355M}  
&N   &          &           &22.13&11.79&12.01&10.07\\
&P.  &          &           &19.26&11.68&11.63&9.91\\
&P.  &\checkmark&           &13.33&9.46&9.00&7.13\\
&P.  &\checkmark&\checkmark &13.33&9.44&8.97&7.13\\
&A.  &          &           &59.50&42.33&47.23&37.00\\
&A.  &\checkmark&           &17.05&18.60&18.54&13.18\\
&A.  &\checkmark&\checkmark &14.37&10.51&9.94&7.81\\
\midrule
\multirow{7}{*}{GPT2-774M}
&N.  &          &           &19.11&10.69&11.29&9.74\\
&P.  &          &           &16.71&10.58&11.08&9.59\\
&P.  &\checkmark&           &11.74&8.56&8.03&6.45\\
&P.  &\checkmark&\checkmark 
&\textbf{11.69}&\textbf{8.52}&\textbf{8.03}&\textbf{6.43}\\
%\cline{2-8}
&A.  &          &           &56.15&42.44&49.25&39.61\\
&A.  &\checkmark&           &14.40&16.83&16.35&11.99\\
&A.  &\checkmark&\checkmark &\textbf{12.13}&\textbf{8.95}&\textbf{8.47}&\textbf{6.78}\\
\midrule
\multirow{7}{*}{GPT2-1.5B} 
&N.  &          &           &17.32&9.87&10.61&9.03\\
&P.  &          &           &15.36&9.81&10.49&9.01\\
&P.  &\checkmark&           &10.84&8.07&7.45&6.01\\
&P.  &\checkmark&\checkmark &\textbf{10.83}&\textbf{8.04}&\textbf{7.43}&\textbf{5.98}\\
%\cline{2-8}
&A.  &          &           &53.30&41.72&49.25&40.25\\
&A.  &\checkmark&           &13.75&16.48&15.20&11.50\\
&A.  &\checkmark&\checkmark 
&\textbf{11.32}&\textbf{8.57}&\textbf{8.01}&\textbf{6.35}\\

\bottomrule
\end{tabular}

\caption{
Perplexity (the lower the better) of OPT models on the entire validation set of WikiText-2 dataset and partial test set of Arxiv, Freelaw and Stackexchange datasets. The retrieval stride is set to 16. \textbf{A.} and \textbf{P.} refers to the appending and prepending context pattern respectively. \textbf{M.T.} refer to the Marking Tokens. And \textbf{N.} refers to none retrieval.
}
\label{tab:app_arxiv_small}
\end{table*}
% \clearpage

\section{Perplexity Results of Llama 3.2}
\label{appendix:e}
We tested the Llama 3.2-3B models in perplexity metrics to demonstrate FlashBack on one of the latest models. The experimental results are presented in table \ref{tab:llama3}.
% \clearpage
\begin{table*}[h!]
\centering
\begin{tabular}{cccccccc}
\toprule
Model &A./P. & LoRA & M.T. & Wikitext-2 & Arxiv & Freelaw & Stackexchange\\
\midrule
\multirow{7}{*}{Llama-3.2-3B}
&N   &          &           &9.02&5.33&6.16&6.14\\
&P.  &          &           &8.20&5.38&6.27&6.12\\
&P.  &\checkmark&           &6.92&4.93&5.95&5.74\\
&P.  &\checkmark&\checkmark &6.93&4.92&5.94&5.74\\
&A.  &          &           &33.00&32.03&38.63&38.06\\
&A.  &\checkmark&           &8.21&6.62&8.42&7.61\\
&A.  &\checkmark&\checkmark &7.56&5.40&6.71&6.25\\
\bottomrule
\end{tabular}
\caption{
Perplexity of Llama-3.2-3B model on the entire validation set of WikiText-2 dataset and
partial test set of Arxiv, Freelaw and Stackexchange datasets. The retrieval stride is set to 16.}
\label{tab:llama3}
\end{table*}

\end{document}